\documentclass[11pt,a4paper]{article}
\usepackage[]{emnlp2020}
\usepackage{times}
\usepackage{latexsym}

\usepackage{microtype}

\usepackage[utf8]{inputenc}
\usepackage{natbib}

\usepackage{tikz}
\usepackage{url}
\usepackage{inconsolata}

\usepackage{amsmath,amsfonts,bm}

\def\eqref#1{equation~\ref{#1}}

\def\1{\bm{1}}

\def\vmu{{\vec{\mu}}}
\def\vnu{{\vec{\nu}}}
\def\vPhi{{\bm{\Phi}}}

\def\neut{{ \mPhi_{\mathrm{ntr}}}}
\def\valpha{{\boldsymbol \alpha}}
\def\vtheta{{\bm{\theta}}}

\def\ve{{\bm{e}}}

\def\vv{{\vec{v}}}
\def\vw{{\vec{w}}}

\def\vz{{\vec{z}}}

\def\mC{{\bm{C}}}

\def\mI{{\bm{I}}}

\def\mK{{\bm{K}}}

\def\mP{{\bm{P}}}

\def\mS{{\bm{S}}}

\def\mV{{\bm{V}}}
\def\mW{{\bm{W}}}

\def\mPhi{{\bm{\Phi}}}

\DeclareMathAlphabet{\mathsfit}{\encodingdefault}{\sfdefault}{m}{sl}
\SetMathAlphabet{\mathsfit}{bold}{\encodingdefault}{\sfdefault}{bx}{n}

\def\calH{{\mathcal{H}}}

\newcommand{\R}{\mathbb{R}}

\let\P\relax
\DeclareMathOperator{\P}{P}

\DeclareMathOperator{\SVD}{SVD}

\newcommand{\kk}{\kappa}
 
\usepackage{amsthm}
\usepackage{amsfonts}
\usepackage{amsmath}
\usepackage{amssymb}
\usepackage{float}
\usepackage{booktabs}
\usepackage{multirow}
\usepackage{cleveref}
\usepackage{adjustbox}
\usepackage{tipa}

\newcommand{\ucambridge}{\normalfont \text{\textipa{D}}}
\newcommand{\ethz}{\text{\normalfont \textipa{Q}}}

\crefname{section}{\S}{\S\S}
\Crefname{section}{\S}{\S\S}
\crefname{table}{Tab.}{}
\crefname{figure}{Fig.}{}
\crefname{algorithm}{Algorithm}{}
\crefname{equation}{Eq.}{}
\crefname{appendix}{App.}{}
\crefname{prop}{Prop}{}
\crefname{method}{Method}{}
\crefformat{section}{\S#2#1#3}  %
\usepackage{adjustbox}
\usepackage[inline]{enumitem}

\usepackage{todonotes}
\makeatletter

\newcommand*\iftodonotes{\if@todonotes@disabled\expandafter\@secondoftwo\else\expandafter\@firstoftwo\fi}  %
\makeatother

\newcommand{\saveForCR}[1]{}

\makeatletter
\newcommand*{\bigcdot}{}%
\DeclareRobustCommand*{\bigcdot}{%
  \mathbin{\mathpalette\bigcdot@{}}%
}
\newcommand*{\bigcdot@scalefactor}{.5}
\newcommand*{\bigcdot@widthfactor}{1.15}
\newcommand*{\bigcdot@}[2]{%
  \sbox0{$#1\vcenter{}$}%
  \sbox2{$#1\cdot\m@th$}%
  \hbox to \bigcdot@widthfactor\wd2{%
    \hfil
    \raise\ht0\hbox{%
      \scalebox{\bigcdot@scalefactor}{%
        \lower\ht0\hbox{$#1\bullet\m@th$}%
      }%
    }%
    \hfil
  }%
}
\makeatother

\aclfinalcopy %
\def\aclpaperid{1018} %

\renewcommand{\R}{\mathbb{R}}
\newcommand{\calD}{\mathcal{D}}
\newcommand{\calE}{\mathcal{E}}
\newcommand{\word}[1]{\textit{#1}}
\renewcommand{\vw}{\boldsymbol{w}}
\renewcommand{\vv}{\boldsymbol{v}}
\renewcommand{\vmu}{{\vec{\mu}}}
\renewcommand{\vnu}{{\boldsymbol \nu}}
\renewcommand{\vz}{\boldsymbol{z}}
\newcommand{\vMu}{{\boldsymbol M}^{\vPhi}}
\renewcommand{\vmu}{{\boldsymbol \mu}}
\newcommand{\defeq}{:=}
\newcommand{\defn}[1]{\textbf{#1}}

\renewcommand{\defeq}{\,{\mathrel{\stackrel{\textnormal{\tiny def}}{=}}}\,}

\makeatletter
\newtheorem*{rep@theorem}{\rep@title}
\newcommand{\newreptheorem}[2]{%
\newenvironment{rep#1}[1]{%
 \def\rep@title{#2 \ref{##1}}%
 \begin{rep@theorem}}%
 {\end{rep@theorem}}}
\makeatother

\newtheorem{theorem}{Theorem}

\newtheorem{proposition}[theorem]{Proposition}
\newtheorem{remark}[theorem]{Remark}
\newtheorem{hypothesis}{Hypothesis}
\newreptheorem{proposition}{Proposition}

\makeatletter
\def\mathcolor#1#{\@mathcolor{#1}}
\def\@mathcolor#1#2#3{%
  \protect\leavevmode
  \begingroup
    \color#1{#2}#3%
  \endgroup
}
\makeatother

\usepackage{xr}
\makeatletter
\newcommand*{\addFileDependency}[1]{%
  \typeout{(#1)}%
  \@addtofilelist{#1}%
  \IfFileExists{#1}{}{\typeout{No file #1.}}%
}
\makeatother

\author{Francisco Vargas$^{\ucambridge}$~\quad~Ryan Cotterell$^{\ucambridge,\ethz}$ \\
  $^{\ucambridge}$University of Cambridge~\quad~%
  $^{\ethz}$ETH Z\"{u}rich \\
  \href{mailto:fav25@cam.ac.uk}{\texttt{fav25@cam.ac.uk}}~\quad~\href{mailto:rcotterell@inf.ethz.ch}{\texttt{rcotterell@inf.ethz.ch}}
}

\title{Exploring the Linear Subspace\\ Hypothesis in Gender Bias Mitigation}
\date{}

\begin{document}

\maketitle

\begin{abstract}
    \newcite{bolukbasi2016man} presents one of the first gender bias mitigation techniques for word representations. Their method takes
    pre-trained word representations as input and attempts to isolate a linear subspace that captures
    most of the gender bias in the representations. As judged by an analogical evaluation task, their method virtually eliminates gender bias in the representations. However, an implicit and untested assumption of their method is that the bias subspace is actually linear. 
    In this work, we generalize their method to a kernelized, non-linear version. We take inspiration from kernel principal component analysis and derive a non-linear bias isolation technique.
    We discuss and overcome some of the practical drawbacks of our method for non-linear gender bias mitigation in word representations and analyze empirically whether the bias subspace is actually linear. Our analysis shows that gender bias is in fact well captured by a linear subspace, justifying
    the assumption of \newcite{bolukbasi2016man}. 
\end{abstract}

\section{Introduction}
Pre-trained word representations are a necessity for strong performance on modern NLP tasks. 
Such representations now serve as input to neural methods \cite{goldberg}, which recently have become the standard models in the field. 
However, because pre-trained representations are constructed from large, human-created corpora, they naturally contain societal biases encoded in that data; gender bias
is among the most well-studied of these biases \cite{caliskan2017semantics}. 
Both a-contextual word representations \cite{mikolov2013efficient, glove} and contextual word representations \cite{elmo, bert} have been shown to encode gender bias \cite{bolukbasi2016man,caliskan2017semantics,zhao2019gender, may2019measuring,karve2019conceptor}. 
More importantly, bias in pre-trained representations has been shown to influence models for downstream tasks where they are used as input, e.g., coreference resolution \citep{CorefRes,CorefRes2}.

\newcite{bolukbasi2016man} present one of the first methods
for detecting and mitigating gender bias in word representations. 
They provide a novel linear-algebraic approach that post-processes
word representations in order to partially remove gender bias. Under their evaluation, they
find they can nearly perfectly remove bias in an analogical reasoning task. 
However, subsequent work \cite{gonen-goldberg-2019-lipstick-pig,hall-maudslay-etal-2019-name} has indicated that gender bias still lingers in the representations, despite \citeposs{bolukbasi2016man} strong experimental results. 
In the development of their method, \newcite{bolukbasi2016man} make a critical and unstated assumption: Gender bias forms a linear subspace of word representation space. 
In mathematics, linearity is a strong assumption and there is no reason \textit{a-priori} why one should expect complex and nuanced societal phenomena, such as gender bias, should be represented by a linear subspace.\looseness=-1

In this work, we present the first non-linear gender bias mitigation technique for a-contextual word representations. In doing so, we directly test the linearity assumption made
by \newcite{bolukbasi2016man}. 
Our method is based on the insight that \citeposs{bolukbasi2016man} method bears a close resemblance to principal component analysis (PCA). Just as one can kernelize PCA \cite{scholkopf1997kernel}, 
we show that one can kernelize the method of \newcite{bolukbasi2016man}. Due to the kernelization,
the bias is removed in the feature space, rather in the word representation space. Thus, we also explore pre-image techniques \cite{mika1999kernel} to project the bias-mitigated vectors
back into $\R^d$. 

As previously noted, there are now multiple bias removal methodologies \citep{CorefRes2,zhao2019gender, may2019measuring} that have succeed the method by \citet{bolukbasi2016man}. Furthermore \citet{gonen-goldberg-2019-lipstick-pig} point out multiple flaws in \citeposs{bolukbasi2016man} bias mitigation technique and the aforementioned methods. 
Nonetheless, we believe that this method has received sufficient attention from the community such that research into its properties is both interesting and useful.

We test our non-linear gender bias technique in several
experiments. First, we consider the Word representation Association Test \cite[WEAT;][]{caliskan2017semantics}; we notice that across five non-linear kernels and convex combinations thereof, there is seemingly no significant difference between
the non-linear bias mitigation technique and the linear one. Secondly, we 
consider the professions task \cite{bolukbasi2016man,gonen-goldberg-2019-lipstick-pig} that measures
how word representations representing different professions are potentially gender-stereotyped. Again, as with the WEAT evaluation, we find that our non-linear bias mitigation technique performs on par with the linear method. We also consider
whether the non-linear gender mitigation technique removes indirect bias 
from the vectors \cite{gonen-goldberg-2019-lipstick-pig}; yet again, we find
the non-linear method performs on par with the linear methods.
As a final evaluation, we evaluate whether non-linear bias mitigation hurts semantic
performance. 
On SimLex-999 \cite{hill2015simlex}, we show that similarity
estimates between the vectors remain on par with the linear methods. 
We conclude that much of the gender bias in word representations is indeed captured by a linear subspace, answering
this paper's titular question. 

\section{Bias as a Linear Subspace}
The first step of \citeposs{bolukbasi2016man}
technique is the discovery of a subspace $B \subset \R^d$ that
captures most of the gender bias. Specifically,
they stipulate that this space is linear.
Given word representations that live in $\R^d$, 
they provide a spectral method for isolating
the bias subspace. 
In this section, we review their approach and show how it is equivalent to principal component analysis (PCA) on a specific design (input) matrix.
Then, we introduce and discuss the implicit assumption made
by their work; we term this assumption the \defn{linear subspace hypothesis} and test it in \cref{sec:kpca}.
\begin{hypothesis}\label{eq:hypothesis}
Gender bias in word representations may be represented
as a linear subspace. 
\end{hypothesis}

\subsection{Construction of a Bias Subspace}\label{sec:construction}
We will assume the existence of a fixed and finite vocabulary $V$, each element of which is a word $\vw_i$. 
The hard-debiasing approach takes a set of {$N$} sets {$\calD = \{D_n\}_{n=1}^N$} as input. Each set {$D_n$}
contains words that are considered roughly semantically equivalent modulo their gender; \newcite{bolukbasi2016man} call the {$D_n$} \defn{defining sets}. For example, {$\{\word{man}, \word{woman}\}$}
and {$\{\word{he}, \word{she}\}$} form two such defining sets. 
We identify each word with a unique integer {$i$} for the sake
of our indexing notation; indeed, we exclusively reserve the index {$i$} for words. 
We additionally introduce the function $f : [|V|] \rightarrow [N]$ that maps an individual word to its
defining set. 
In general, 
the defining sets are not limited to a cardinality of two, but in practice \citet{bolukbasi2016man} exclusively employ defining sets
with a cardinality of two in their experiments. 
Using the sets $D_n$, \citet{bolukbasi2016man} define the matrix $\mC$\looseness=-1
\begin{equation}
\begin{aligned}
\mC \defeq \sum_{n=1}^N \frac{1}{|D_n|} \sum_{i \in D_n}(\vw_i - \vmu_{n})(\vw_i - \vmu_{n})^{\top},
\end{aligned}
\end{equation}
where we write {$\vw_i$} for the {$i^\text{th}$} word's representation and the empirical mean vector $\vmu_{n}$ is defined as
\begin{equation}
    \vmu_{n} \defeq \frac{1}{|D_n|} \sum_{i\in D_n} \vw_i. 
\end{equation}
\newcite{bolukbasi2016man} then extract a bias subspace {$B$} using the singular value decomposition (SVD). 
Specifically, they define the bias subspace to be the space
spanned by the first $k$ columns of $\mV$ where
\begin{equation}\label{eq:svd}
\mV \mS \mV^{\top} = \SVD(\mC).
\end{equation} 
As $\mC$ is symmetric and positive semi-definite,
the SVD is equivalent to an eigendecomposition as our notation in \cref{eq:svd} shows. 
We assume the columns of $\mV$, the eigenvectors of $\mC$, are ordered by the magnitude of their corresponding eigenvalues.\looseness=-1

\subsection{Bias Subspace Construction as PCA}\label{sec:pca}
As briefly noted by \newcite{bolukbasi2016man}, this can thus be cast as performing principal component analysis (PCA) on a recentered input matrix. We prove this assertion
more formally. We first prove
that the matrix {$\mC$} may be written
as an empirical covariance matrix.
\begin{proposition}\label{prop1}
Suppose {$|D_n| = 2$} for all {$n$}. Then we have
\begin{equation}
    \mC = \frac{1}{2} \sum_{i=1}^{2N} \mW_{i,\bigcdot }^{\top}\,\mW_{i,\bigcdot},
\end{equation}
where we define the design matrix {$\mW$} as:
\begin{align}\label{eq:design-matrix}
\mW_{i,\bigcdot} =\left( \vw_i - \vmu_{f(i)} \right)^{\top}. 
\end{align}
\end{proposition}
\begin{proof}
\begin{subequations}
\begin{align}
      \mC &= \sum_{n=1}^N \frac{1}{|D_n|} \sum_{i \in D_n}\left( \vw_i - \vmu_{n} \right)\left(\vw_i - \vmu_{n} \right)^{\top} \nonumber  \\
     &= \frac{1}{2} \sum_{n=1}^N  \sum_{i \in D_n}\left( \vw_i - \vmu_{n} \right)\left(\vw_i - \vmu_{n} \right)^{\top} \\
     &= \frac{1}{2} \sum_{i=1}^{2N} \left( \vw_i - \vmu_{f(i)} \right)\left(\vw_i - \vmu_{f(i)} \right)^{\top} \\
     &=\frac{1}{2} \sum_{i=1}^{2N} \mW_{i,\bigcdot}^{\top}\,\mW_{i,\bigcdot},
\end{align}
\end{subequations}
where $\mW_{i,\bigcdot} \in \mathbb{R}^{2N \times d}$ is defined as above.
\end{proof}
Next, we show that the matrix is centered, which is a
requirement for PCA.
\begin{proposition}\label{prop2}
The matrix {$\mW$} is row-wise centered.
\end{proposition}
\begin{proof}
\begin{subequations}
\begin{align}
    \frac{1}{2} \sum_{i=1}^{2N} \mW_{i,\bigcdot}^{\top} &= \frac{1}{2} \sum_{i=1}^{2N} \left( \vw_i - \vmu_{f(i)} \right) \\
    &=\frac{1}{2} \sum_{n=1}^{N} \sum_{i \in \calD_n} \left( \vw_{i} - \vmu_{n} \right) \\
     &=\frac{1}{2} \sum_{n=1}^{N} \left( - 2 \vmu_{n} + \sum_{i \in \calD_n} \vw_{i}  \right) \\
       &=\frac{1}{2} \sum_{n=1}^{N} \left( 2 \vmu_{n} - 2 \vmu_{n} \right) = 0.
\end{align}
\end{subequations}
\end{proof}
\begin{remark}
The method of \newcite{bolukbasi2016man} may be considered principal component analysis performed on the matrix {$2 \mC$}.
\end{remark}
\begin{proof}
As the algebra in \Cref{prop1} and \Cref{prop2}
show we may formulate the problem as an SVD on
a mean-centered covariance matrix. One view
of PCA is performing matrix factorization on such
a matrix.
\end{proof}

\section{\newcite{bolukbasi2016man}}
In this section, we review the bias mitigation technique introduced by \newcite{bolukbasi2016man}.
When possible, we take care to reformulate their method in terms of
matrix notation. 
They introduce a two-step process that \defn{neutralizes} and \defn{equalizes}
the vectors to mitigate gender bias in the representations. 
The underlying assumption of their method is that there exists
a linear subspace $B \subset \R^d$ that captures most of the
gender bias.

\subsection{Neutralize\label{sub:neut}}
After finding the linear bias subspace $B$,
the gist behind \citeposs{bolukbasi2016man} approach is
based on elementary linear algebra. We may decompose
any word vector {$\vw$} as the sum of its orthogonal projection onto the bias subspace (range of the projection) and its orthogonal projection onto the complement of the bias subspace (null space of the projection), i.e.,
\begin{equation}
    \vw = \vw_{B} + \vw_{\perp B}.
\end{equation}
We may then re-embed every vector as 
\begin{equation}
    \vw_{\mathrm{ntr}} \defeq \vw - \vw_B = \vw_{\perp B}.
\end{equation}
We may re-write this in terms of matrix notation in
the following manner. Let $\{\vv_k\}_{k=1}^K$ be an orthogonal basis for the linear bias subspace {$B$}. 
This may be found by taking the eigenvectors {$\mC$} that correspond to the top-$K$ eigenvalues with largest magnitude.
Then, we define the projection matrix onto the bias subspace
as $\mP_K = \sum_{k=1}^K \vv_k \vv_k^{\top}$ it follows that
the matrix $(\mI - \mP_K)$ is a projection matrix on the complement of $B$. 
We can then write the neutralize step using matrices
\begin{equation}
    \vw_{\mathrm{ntr}} \defeq (\mI - \mP_K)\vw. \label{neutralise}
\end{equation}
The matrix formulation of the neutralize step
offers a cleaner presentation of what the neutralize
step does: it projects the vectors onto the orthogonal
complement of the bias subspace.

\subsection{Equalize}
\newcite{bolukbasi2016man} decompose words into two classes. 
The neutral words which undergo neutralization as explained above, and the gendered words, some of which receive the equalizing treatment. Given a set of \defn{equality sets} {$\calE = \{E_1, \ldots, E_L\}$} which we can see as a greater extension of the defining sets {$\calD$}, i.e., {$E_i =\{\word{guy}, \word{gal}\}$}, \citet{bolukbasi2016man} then proceed to decompose each of the words  {$\vw \in \calE$} into their gendered and neutral counterparts, setting their neutral component to a constant (the mean of the equality set) and the gendered component to its mean-centered projection into the gendered subspace:
\begin{equation}
    \vw_{\mathrm{eq}} \defeq \vnu + Z\,(\vw_{B} - \vmu_{B}) \label{amon}
\end{equation}
where we define the following quantities:
\begin{subequations}
\begin{align}
    Z &\defeq \frac{\sqrt{1 - ||\vnu||_2^2}}{|| \vw_{B} - \vmu_{B} ||_2} \\
    \vmu &\defeq \frac{1}{|E_i|} \sum_{\vw \in E_i} \vw  \\
    \vnu  &\defeq (\mI - \mP_K)\vmu,
\end{align}
\end{subequations}
the ``normalizer'' {$Z$} ensures the vector is of unit length.
This fact is left unexplained in the original work,
but \newcite{hall-maudslay-etal-2019-name} provide
a proof in their appendix.

\section{Bias as a Non-Linear Subspace \label{sec:kpca}}
We generalize the framework presented in \citet{bolukbasi2016man} and cast it to a non-linear setting by exploiting its relationship to PCA. Thus, the natural extension of \citet{bolukbasi2016man} is to kernelize it analogously to \citet{scholkopf1997kernel}, which is the kernelized generalization of PCA. 
Our approach preserves all the desirable formal properties presented in the linear method of \newcite{bolukbasi2016man}.\looseness=-1

\subsection{Adapting the Design Matrix}
The idea behind our non-linear bias mitigation technique is based on kernel PCA \cite{scholkopf1998nonlinear}. In short, the idea is to map the original word representations {$\vw_i \in \R^{d}$} to a higher-dimensional space {$\calH$} via a function {$\vPhi: \mathbb{R}^{d} \rightarrow \calH$}. We will consider cases
where {$\calH$} is a reproducing kernel Hilbert space (RKHS) with reproducing kernel {$\kk(\vw_{i}, \vw_{j}) = \langle \vPhi(\vw_{i}) , \vPhi(\vw_{j}) \rangle$} where the notation {$\langle \cdot, \cdot \rangle$} refers to an inner product in the RKHS. Traditionally,
one calls {$\calH$} the feature space and will use this terminology
throughout this work. Exploiting the reproducing kernel property, we may carry out 
\citeposs{bolukbasi2016man} bias isolation technique and construct a non-linear analogue.

We start the development of bias mitigation technique in feature space with a modification of the design matrix presented in \cref{eq:design-matrix}.
In the RKHS setting the non-linear analogue is
\begin{align}
    \mW^{\vPhi}_{i, \bigcdot} := \vPhi(\vw_i) - \vMu_{D_f(i)}, \quad \forall\,\vw_i \in V,
\end{align}
where we define
\begin{align}
     \vMu_{D_n} := \frac{1}{|D_n|}\sum_{i \in D_n}\vPhi(\vw_i).
\end{align}
As one can see, this is a relatively straightforward mapping
from the set of linear operations to non-linear ones.\looseness=-1

\subsection{Kernel PCA}
Using our modified design matrix, we can 
cast our non-linear bias mitigation technique
as a form of kernel PCA. Specifically,
we form the matrix
\begin{equation}
    \mC^{\vPhi} = \frac{1}{2} \sum_{i=1}^{2N} (\mW^{\vPhi}_{i,\bigcdot})^{\top} \mW^{\vPhi}
    _{i, \bigcdot}.
\end{equation}
 Our goal is to find the eigenvalues {$\lambda_k$} and their corresponding eigenfunctions {$\mV_k^{\vPhi} \in \calH$} by solving the eigenvalue problem
\begin{align}\label{eq:eigen1}
    \mC^{\vPhi}\,\mV^{\vPhi}_k &= \lambda_k \mV^{\vPhi}_k.
\end{align}
Computing these directly from \cref{eq:eigen1}
is impossible since $\calH$'s dimension may 
be prohibitively large or even infinite.
However, \citeauthor{scholkopf1997kernel} note that {$\mV_k^{\vPhi}$} is spanned by {$\{\vPhi(\vw_i)\}_{i=1}^{2N}$}. 
This allows us to rewrite \cref{eq:eigen1} as
\begin{align}\label{eq:span}
    \mV^{\vPhi}_k  = \sum_{i=1}^{2N} \alpha^{k}_{i}\,\vPhi(\vw_i),
\end{align}
where there exist coefficients {$ \alpha^{k}_i \in \mathbb{R}$}.
Now, by substituting \cref{eq:span} and \cref{eq:eigen1} into the respective terms in {$\lambda \langle \vPhi(\vw_i),\mV_k^{\vPhi} \rangle = \langle \vPhi(\vw_i),\mC\,\mV_k^{\vPhi} \rangle$}, \citet{scholkopf1997kernel} derive a computationally feasible eigendecomposition problem. Specifically, they consider
\begin{align}
    2N \lambda_k\,\valpha^k = \mK \valpha^k, \label{eigk}
\end{align}
where {$\mK_{ij} = \langle \vPhi(\vw_i), \vPhi(\vw_j) \rangle$}. 
Once all the $\valpha^k$ vectors have been estimated the inner product between an eigenfunction $\mV_k^{\vPhi}$ and a point $\vw$ can be computed as
\begin{subequations}
\begin{align}
    \beta_k(\vw) &= \langle \mV^{\vPhi}_k , \vPhi(\vw_i) \rangle \\
    &= \sum_{i=1}^{2N} \alpha_i^k\,\kk(\vw_i, \vw)  \label{beta}
\end{align}
\end{subequations}
A projection into the basis {$\{\mV^{\vPhi}_k\}_{k=1}^K$} can then be carried out by applying the projection operator {$\P_K : \calH \rightarrow \calH$} as follows:
\begin{align}
    \P_K \vPhi(\vw) = \sum_{k=1}^K\beta_k(\vw)\,\mV^{\vPhi}_k, \label{proj}
\end{align}
where $K$ is the number of principal components. Projection operator {$\P_K$} is analogous to the linear projection {$\mP_K$} introduced in \cref{sub:neut}. 
\subsection{Centering Kernel Matrix}
We can perform the required mean-centering operations on the design matrix by centering the kernel matrix in a similar fashion to \citet{scholkopf1998nonlinear}.  For the case of equality sets of size 2, which is what  \citeauthor{bolukbasi2016man} use in practice, we realize that the centered design matrix reduces to pairwise differences:
\begin{equation}
\begin{aligned}
    \mW^{\vPhi}_{i, \bigcdot} &= \frac{1}{2} (\vPhi(\vw_i) -\vPhi(\vw_j)),  \\
    &\quad\quad\quad \vw_i, \vw_j \in D_n \wedge \vw_j \neq \vw_i 
\end{aligned}
\end{equation}
which leads to a very simple re-centering in terms of the Gram matrices:
\begin{equation}
\begin{aligned}
    \widetilde{\mK} = \mK^{(11)} - \mK^{(12)} - (\mK^{(12)})^{\top} + \mK^{(22)}
\end{aligned}
\end{equation}
where 
\begin{subequations}
\begin{align}
    \mK^{(xy)}_{ij} &= \frac{1}{2}\kk\left(\mW^{(x)}_{i,\bigcdot}, \mW^{(y)}_{j, \bigcdot}\right),  \\
    \mW^{(z)}_{i.\bigcdot} &= \begin{cases}
    \vw_{\pi_{1 + i\text{\,mod\,}2}(\hat{f}({\lceil \frac{i}{2} \rceil}))} & z = 1 \\
    \vw_{\pi_{2 -i\text{\,mod\,}2}(\hat{f}({\lceil \frac{i}{2} \rceil}))} & z = 2
    \end{cases}\label{eq:checkerboard}
\end{align}
\end{subequations}
where {$\hat{f}: [N] \rightarrow [|V|]\times[|V|]$} maps a defining set index to a tuple containing the word indices in the corresponding defining set and {$\pi_1,\pi_2:  [|V|]\times[|V|] \rightarrow [|V|]$} are projection operators which return the first or second elements of a tuple respectively. In simpler terms, \cref{eq:checkerboard} is creating two matrices: matrix {$\mW^{(1)}$} which is constructed by looping over the definition sets and placing pairs within the same definition set as adjacent rows, then {$\mW^{(2)}$} is constructed in the same way but the order of the adjacent pairs is swapped relative to {$\mW^{(1)}$}.
Once we have this pairwise centered Gram matrix {$\widetilde{\mK}$} we can apply the eigendecomposition procedure described in \cref{eigk} directly on {$\widetilde{\mK}$}. We note that carrying out this procedure using a linear kernel recovers the linear bias subspace from \citet{bolukbasi2016man}.\looseness=-1

\subsection{Inner Product Correction (Neutralize) \label{sub:neutkpca}}
We now focus on neutralizing and equalizing the inner products in the RKHS rather than correcting the word representations directly. 
Just as in the linear case, we can decompose the representation of a word in the RKHS into biased and neutral components
\begin{align}
    \mPhi(\vw) = \P_K^{\perp}\mPhi(\vw) + \P_K \mPhi(\vw),
\end{align}
which provides a nonlinear equivalent for \cref{neutralise}:
\begin{equation}
\begin{aligned}
    \neut(\vw) &= \P_K^{\perp}\mPhi(\vw)  \\
    &= \mPhi(\vw) - \P_K \mPhi(\vw).
\end{aligned}
\end{equation}
\begin{proposition}\label{prop:neut}
The corrected inner product in the feature space for two neutralized words {$\vz, \vw$} is given by 
\begin{equation}
\begin{aligned}
   \langle \neut&(\vz),\neut(\vw) \rangle \label{eq:neutrkhs} \\
   &=\kk(\vz, \vw) - \sum_{k=1}^{K}\beta_k(\vz)\,\beta_k(\vw)
\end{aligned}
\end{equation}
\end{proposition}
\begin{proof}
\begin{equation}
\begin{aligned}
    \Big<&\neut(\vz), \neut(\vw) \Big> = \\
    &\Big<\vPhi(\vz)- \P_K \vPhi(\vz),\: \vPhi(\vw)- \P_K \vPhi(\vw)\Big> 
\end{aligned}
\end{equation}
Applying \cref{beta} and \cref{proj}
\begin{subequations}
\begin{align}
    &=\kk(\vz, \vw) - \sum_{k=1}^{K}\beta_k(\vw)\sum_{i=1}^{2N}\alpha_i^{k} \kk(\vz, \vw_{i})  \\ &- \sum_{k=1}^{K}\beta_k(\vz)\sum_{i=1}^{2N}\alpha_i^{k} \kk(\vw, \vw_{i})+\sum_{k=1}^{K}\beta_k(\vw)\beta_k(\vz)  \nonumber\\
    &= \kk(\vz, \vw) - \sum_{k=1}^{K}\beta_k(\vw)\sum_{i=1}^{2N}\alpha_n^{k} \kk(\vz, \vw_{i}) \\
    &= \kk(\vz, \vw) - \sum_{k=1}^{K}\beta_k(\vw)\beta_k(\vz), \label{eq:testsim1}
\end{align}
\end{subequations}
 where 
 \begin{equation}
 \beta_k(\vw)=\bm{V}_{k}^{\vPhi\top} \vPhi(\vw) \sum_{i=1}^{2N}\alpha_{i}^{k}\,\kk(\vw, \vw_i), 
\end{equation}
 as derived by \citet{scholkopf1998nonlinear}. 
\end{proof}
An advantage of this approach is that it will not rely on errors due to the approximation of the pre-image. 
However, it will not give us back a set of bias-mitigated representations. Instead, it returns a bias mitigation metric, thus limiting the classifiers and regressors we could use.  
\Cref{eq:neutrkhs} provides us with an {$\mathcal{O}(KNd)$} approach to compute the inner product between words in feature space.\looseness=-1

\subsection{Inner Product Correction (Equalize)}
To equalize, we may naturally convert \cref{amon} to its feature-space equivalent.
We define an equalizing function\looseness=-1
\begin{equation}
    \vtheta(\vw_i) \defeq \vnu^{\vPhi}+ Z^{\vPhi}\,\P_K\left(\vPhi(\vw_i) - \vMu_{E_{g(i)}}\right) 
\end{equation}
where we define
\begin{subequations}
\begin{align}
     \vnu^{\vPhi} &\defeq \P_K^{\perp} \vMu_{E_{g(i)}}  \\
     Z^{\vPhi} & \defeq \frac{\sqrt{1- || \nu_{\vPhi}||^2}}{|| \P_K(\vPhi(\vw) - \vMu_{E_{g(i)}}) ||} 
\end{align}
\end{subequations}
where {$g: [|V|] \rightarrow [L]$} maps an individual word index to its corresponding equality set index. Given vector dot products in the linear case follow the same geometric properties as inner products in the RKHS we can show that {$\vtheta_{\mathrm{eq}}(\vw)$} is unit norm follows directly from the proof for Proposition 1 in \cite{hall-maudslay-etal-2019-name} which can be found in Appendix A of \citet{hall-maudslay-etal-2019-name}.
\begin{proposition}
For any two vectors in the observed space {$\vw, \vz$} and their corresponding representations in feature space {$\vPhi(\vw), \vPhi(\vz)$} the inner product {$\langle \vPhi(\vw) - \P_K\vPhi(\vw), \P_K\vPhi(\vz) \rangle$} is {$0$}. \label{prep0}
\end{proposition}
\begin{proof}
\begin{align*}
    \langle \vPhi&(\vw) - \P_K\vPhi(\vw), \P_K\vPhi(\ve)\rangle \\ \nonumber\\ &=\langle\vPhi(\vw),\P_K\vPhi(\vz)\rangle  - \langle \P_K\vPhi(\vw),\P_K\vPhi(\vz) \rangle \nonumber \\
    &= \sum_{k=1}^{K}\beta_k(\vw)\beta_k(\vz) - \sum_{k=1}^{K}\beta_k(\vw)\beta_k(\vz) = 0.\nonumber
\end{align*}
\end{proof}
\begin{proposition}\label{prop:amon}
For a given neutral word {$\vw$} and a word in an equality set {$\ve \in \calE$} the inner product {$\left<\neut(\vw), \vtheta(\ve)\right>$} is invariant across members in the equality set {$\calE$}.
\end{proposition}
\begin{proof}
\begin{align*}
    \langle & \neut(\vw), \vtheta(\ve) \rangle\overset{\textit(i)}{=} \langle \neut(\vw),  \vnu^{\vPhi} \rangle \\
     &=  \frac{1}{|E|} \sum_{i \in E} \langle\neut(\vw),  \P_K^{\perp}\vPhi(\vw_i) \rangle \nonumber \\
     &=  \frac{1}{|E|} \sum_{i \in E} \langle \P_K^{\perp}\vPhi(\vw),  \P_K^{\perp}\vPhi(\vw_i) \rangle \\
     &\overset{\textit(ii)}{=}  \frac{1}{|E|} \sum_{i \in E}\left(\kk(\vw, \vw_i) - \sum_{k=1}^K \beta_{k}(\vw)\beta_{k}(\vw_i) \right),
\end{align*}
where \textit{(i)} follows from \Cref{prep0} and \textit{(ii)} follows from \Cref{prop:neut}. 
\end{proof}
At this point, we have completely kernelized the approach in \citet{bolukbasi2016man}. Note that a linear kernel reduces to the method described in \citet{bolukbasi2016man} as we would expect. We can see an initial disadvantage that equalizing via inner product correction has in comparison to \citet{bolukbasi2016man} and that is that we now require switching in between three different inner products at test time depending on whether the words are neutral or not. To overcome this in practice, we neutralize all words and do not use the equalize correction, however, we present it for completeness.
\begin{figure}
    \centering
    \includegraphics[width=\linewidth]{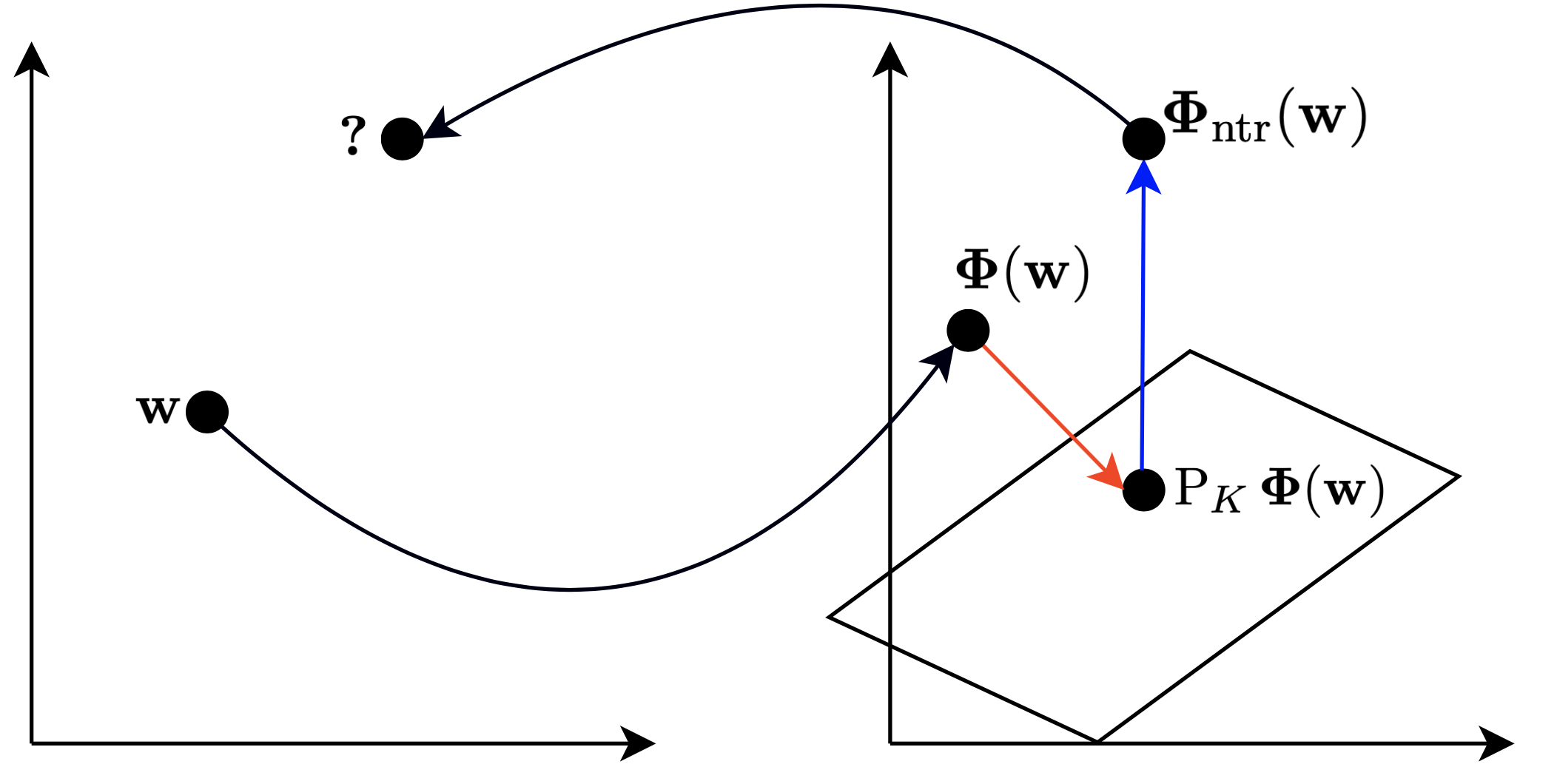}
    \caption{Pre-image problem illustration for the neutralized representations (null-space). 
    The plane represents the bias subspace in the RKHS.}
    \label{fig:my_rkhs}
\end{figure}
\section{Computing the Pre-Image}\label{sub:preimage}
As mentioned in the previous section, a downfall of the metric correction approach is that it does not provide us representations that we can use in downstream tasks: the bias-mitigated representations only exist in feature space. 
Thus, when it comes to transfer tasks such as classification we are limited to kernel methods, i.e., support vector machines). 
One way to resolve this problem is by obtaining the pre-image of the corrected representations in the feature space. 

Finding the pre-image is a well-studied problem for kernel PCA \cite{kwok2004pre}. The goal is to fine the pre-image mappings {$\Gamma: \mathcal{H}_K \oplus  \mathcal{H}_{K}^{\perp} \rightarrow \R^d$}, {$\Gamma^{\perp}: \mathcal{H}_{K}^{\perp} \rightarrow \R^d$} and {$\Gamma^{\top}: \mathcal{H}_{K} \rightarrow \R^d$} that compute (or approximate) the pre-images for {$\vPhi(\vw_i), \vPhi_{\P_K}^{\perp}(\vw_i)$} and {$\vPhi_{\P_K}(\vw_i)$}, respectively. In our case,
with the pre-image mapping, the neutralize step from \citet{bolukbasi2016man} becomes
\begin{align}
    \vz_i^\star = \Gamma^{\perp}\left(\vPhi(\vw_i) - \P_K\vPhi(\vw_i) \right).
\end{align}
In general, we will not have access to $\Gamma^{\perp}$
so we fall back on the following approximation scheme.

\paragraph{Additive Decomposition Approach.}
Alternatively, following \citet{kandasamy2016additive}, we can construct an approximation to $\Gamma$ that additively
decomposes over the direct sum $\oplus$. 
We decompose $\Gamma$ additively over the direct sum $\mathcal{H}_K \oplus  \mathcal{H}_{K}^{\perp}$. 
That is, we assume that the pre-image mappings have the following form:
\begin{table}
\begin{tabular}{@{}ll@{}}
\toprule
\textbf{kernel}             & {$\kk(\vw_i, \vw_j)$}                                                                                                                                                         \\ \midrule
Cosine             & {$\frac{\vw_i^{\top} \vw_j}{||\vw_i||_2\,||\vw_j||_2}$}                                                                                           \\
RBF Kernel         & {$\exp\left(-\gamma\,|| \vw_i - \vw_j||_2^2\right)$}                                                                                                      \\
Sigmoid Kernel     & {$\tanh(\gamma\,\vw_i^{\top} \vw_j + c_0)$}                                                                                                  \\
Polynomial Kernel  & {$(\gamma\,\vw_i^{\top} \vw_j + c_0)^d$}                                                                                                     \\
Laplace Kernel     & {$ \exp(-\gamma\,|| \vw_i - \vw_j||_1)$}                                                                                                     \\
Convex Combination & {$\sum_{\ell=1}^L  \alpha_\ell \,\kk_n(\vw_i, \vw_j),$} \\ \bottomrule
\end{tabular}
\caption{Different kernels used throughout experiments.} \label{tab:kern}
\end{table}
 \begin{table*}[]
\begin{tabular}{@{}p{2.5cm}llclllllll@{}}
\toprule
\multirow{2}{*}{ Targets } & \multicolumn{2}{l}{Original} & \multicolumn{2}{l}{PCA} & \multicolumn{2}{l}{KPCA (rbf)} & \multicolumn{2}{l}{KPCA (sig)} & \multicolumn{2}{l}{KPCPA (lap)} \\ \cmidrule(l){2-11} 
                         & d                 & p                 & d               & p              & d               & p              & d               & p              & d               & p              \\ \midrule
\multicolumn{11}{l}{    \hspace{3.5cm} Google News Word2Vec \cite{mikolov2013efficient} } \\
\midrule 
Career , Family        & 1.622             & 0.000             & 1.327           & 0.001          & 1.321           & 0.005          & 1.319           & 0.006          & 1.311           & 0.002          \\
Math, Arts         & 0.998             & 0.017             & -0.540          & 0.859          & -0.755          & 0.922          & -0.754          & 0.933          & -0.024          & 0.507          \\
Science , Arts            & 1.159             & 0.005             & 0.288           & 0.281          & 0.271           & 0.307          & 0.269           & 0.283          & 1.110           & 0.009          \\ \midrule

\multicolumn{11}{l}{    \hspace{5cm} GloVe \cite{glove}} \\
 \midrule
Career , Family          & 1.749         & 0.000        & 1.160       & 0.007     & 1.166          & 0.006        & 1.165          & 0.01         & 1.443          & 0.000        \\
Math, Arts               & 1.162         & 0.007        & 0.144       & 0.389     & 0.096          & 0.437        & 0.095          & 0.411        & 0.999          & 0.015        \\
Science , Arts           & 1.281         & 0.008        & -1.074      & 0.985     & -1.114         & 0.995        & -1.112         & 0.993        & -0.522         & 0.839        \\ \bottomrule
\end{tabular}
\caption{ WEAT results using GloVe and Google News word representations.} \label{tab:weat_glove}
\end{table*}
\begin{subequations}
\begin{align}
    &\Gamma(\mPhi(\vw)) = \\ &\quad\quad\quad\ \Gamma^{\perp}\left(\mPhi_{\P_K}^{\perp}(\vw) \right) + \Gamma^{\top}(\mPhi_{\P_K}(\vw) ) \nonumber \\
     &\Gamma^{\perp}\left(\mPhi_{\P_K}^{\perp}(\vw)\right)   =\\ 
   &\quad\quad\quad\quad \Gamma(\mPhi(\vw))  - \Gamma^{\top}(\mPhi_{\P_K}(\vw) ) \nonumber 
\end{align}
\end{subequations}
Given that we will always know that the pre-image of {$\Gamma(\mPhi(\vw))$} exists and is {$\vw$}, we can select {$\Gamma$} to return {$\vw$} resulting in:
{\begin{align}
    \Gamma^{\perp}\left(\mPhi_{\P_k}^{\perp}(\vw) \right) = \vw - \Gamma^{\top}(\mPhi^{\top}(\vw) ) \label{preimageq}
\end{align}}
We then learn an analytic approximation for {$\Gamma^{\top}$} using the method in \citet{bakir2004learning}.
\begin{table}
\begin{adjustbox}{width=\columnwidth}
\begin{tabular}{@{}lllll@{}}
\toprule
\multirow{2}{*}{Dataset} & \multicolumn{2}{l}{\newcite{bolukbasi2016man}} & \multicolumn{2}{l}{PCA}            \\ \cmidrule(l){2-5} 
                         & d              & p            & \multicolumn{1}{c}{d}      & p     \\ \midrule
\multicolumn{5}{l}{Google News Word2Vec \cite{mikolov2013efficient}}                                                      \\ \midrule
Career , Family          & 1.299          & 0.003        & 1.327                      & 0.001 \\
Math, Arts               & -1.173         & 0.995        & -0.540                     & 0.859 \\
Science , Arts           & -0.509         & 0.832        & 0.288                      & 0.281 \\ \midrule
\multicolumn{5}{l}{    \hspace{1cm} GloVe \cite{glove}}                                                                     \\ \midrule
Career , Family          & 1.160          & 0.000        & \multicolumn{1}{c}{1.160}  & 0.007 \\
Math, Arts               & -0.632         & 0.887        & \multicolumn{1}{c}{0.144}  & 0.389 \\
Science , Arts           & 0.937          & 0.937        & \multicolumn{1}{c}{-1.074} & 0.985 \\ \bottomrule
\end{tabular}
\end{adjustbox}
\caption{Effect of the equalize step} \label{equalizetab}
\vspace{-10pt}
\end{table}
Note that most pre-imaging methods, e.g., \citet{mika1999kernel} and \newcite{bakir2004learning}, are designed to approximate a pre-image {$\Gamma^{\top}$} and do not generalize to approximating the pre-image mappings {$\Gamma$} and {$\Gamma^{\perp}$}. 
This is because such methods explicitly optimize for pre-imaging representations in the space {$\Gamma^{\top}$} using points in the training set as examples of their pre-image, for the null-space {$\Gamma^{\perp}$} we have no such examples.\looseness=-1

\section{Experiments and Results} \label{sec:experiments}
We carry out experiments across a range of benchmarks and statistical tests designed to quantify the underlying bias in word representations \cite{gonen-goldberg-2019-lipstick-pig}. Our experiments focus on quantifying both direct and indirect bias as defined in \citet{gonen-goldberg-2019-lipstick-pig,hall-maudslay-etal-2019-name}. 
Furthermore, we also carry out word similarity experiments using the \citet{hill2015simlex} benchmark in order to assess that our new bias-mitigated spaces still preserve the original properties of word representations \cite{mikolov2013efficient}.

\subsection{Experimental Setup}
Across all experiments we apply the neutralize metric correction step to all word representations, in contrast to \citet{bolukbasi2016man} where the equalize step is applied to the equality sets {$\calE$} and the neutralize step to a set of neutral words as determined in \citet{bolukbasi2016man}.
We show in \cref{equalizetab} that applying the equalize step does not bring an enhancement over neutralizing all words. We varied kernel hyper-parameters using a grid search and found that they had little effect on performance, as a result we used default initialization strategies as suggested in \citet{scholkopf1998nonlinear}. Unless mentioned otherwise, all experiments use the inner product correction approach introduced in \cref{sub:neutkpca}.

\begin{figure}
    \hspace{-0.5cm}\includegraphics[scale=0.26]{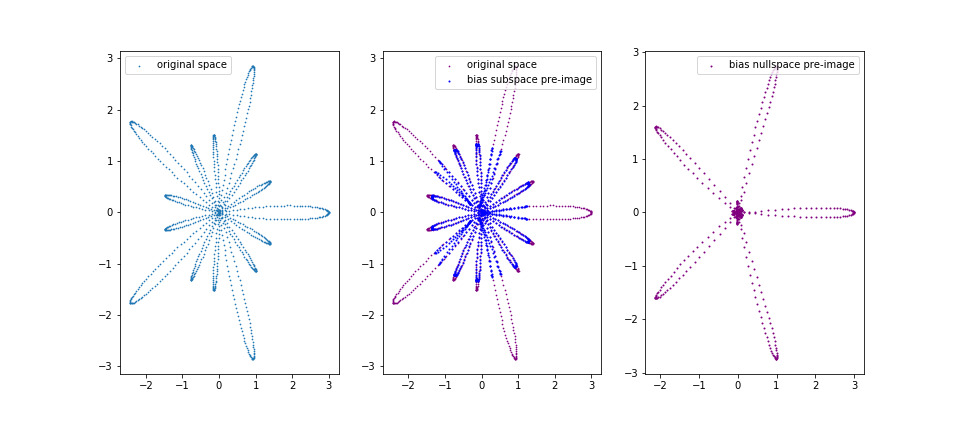}
    \caption{2D toy example of non-linear component removal using Kernel PCA and the pre-image (neutralize step) described in \cref{sub:preimage}.}
    \label{fig:removal}
\end{figure}

\subsection{Kernel Variations}

The main kernels used throughout experiments are specified in \cref{tab:kern}. We also explored the following compound kernels:
\begin{enumerate*}[label=(\roman*)]
    \item convex combinations of the Laplace, radial basis function (RBF), cosine and sigmoid kernels;
    \item convex combinations of cosine similarity, RBF, and sigmoid kernels;
    \item convex combinations of RBF and sigmoid kernels;
    \item polynomial kernels up to 4$^{\text{th}}$ degree.
\end{enumerate*}
We only report the results on the most fundamental kernels out of the explored kernels.

\subsection{Direct Bias: WEAT}
The Word Embedding Association Test \newcite[WEAT;][]{caliskan2017semantics} is a statistical test analogous to the implicit association test (IAT) for quantifying human biases in textual data \citep{greenwald1995implicit}. 
WEAT computes the difference in relative cosine similarity between two sets of target words {$X$} and {$Y$} (e.g., careers and family) and two sets of attribute words {$A$} and {$B$} (e.g., male names and female names). Formally, this quantity is Cohen's {$d$}-measure \cite{cohen1992statistical} also known as the effect size: The higher the measure, the more biased the representations. To quantify the significance of the estimated {$d$}, \newcite{caliskan2017semantics} define the null hypothesis that there is no difference between the two sets of target words and the sets of attribute words in terms of their relative similarities (i.e., {$d=0$}). Using this null hypothesis, \citet{caliskan2017semantics} then carry out a one-sided hypothesis test where failure to reject the null-hypothesis {$(p > 0.05)$} means that the degree of bias measured by {$d$} is not significant.

 We obtain WEAT scores across different kernels (\cref{tab:weat_glove}). We observe that the differences between the linear and the non-linear kernels is small and, in most cases, the linear kernel has a smaller value for the effect size indicating a lesser degree of bias in the corrected space. Overall, we conclude that the non-linear kernels do not reduce the linear bias as measured by WEAT further than the linear kernels. We also experiment with polynomial kernels and obtain similar results, which can be found in \cref{tab:polyglove} of \cref{apdx:poly}.\looseness=-1
 
\subsection{Professions \cite{gonen-goldberg-2019-lipstick-pig}}\label{subsec:professions}

We consider the professions dataset introduced by \newcite{bolukbasi2016man} and apply the benchmark defined in \citet{gonen-goldberg-2019-lipstick-pig}. We find the neighbors (100 nearest neighbors) of each word using the corrected cosine similarity and count the number of male neighbors. We then report the Pearson correlation coefficient between the number of male neighbors for each word and the original bias of that word. The original bias of a word vector {$\vw$} is given by the cosine similarity {$\cos(\vw, \textbf{\textit{he}} - \textbf{\textit{she}})$} in the original word representation space. 
We can observe from the results in \cref{professions} that the non-linear kernels yield only marginally different results, which in most cases seem to be slightly worse, i.e., their induced space exhibits marginally higher correlations with the original biased vector space.

\begin{table}
\centering
\begin{adjustbox}{width=\columnwidth}
\begin{tabular}{@{}llllll@{}}
\toprule
representations & Original & PCA   & KPCA(rbf) & KPCA(sig) & KPCA(lap) \\ \midrule
Word2Vec   & 0.740    & 0.675 & 0.678     & 0.675     &   0.708        \\
Glove      &     0.758     &  0.675     & 0.681          &    0.680       &       0.715    \\ \bottomrule
\end{tabular}
\end{adjustbox}
\caption{Pearson correlation coefficients of professions analogy task. All observed at significant at {$\alpha= 0.05$}. Indeed,
all have {$p$}-values  {$<10^{-30}$}.} \label{professions}
\end{table}

 \begin{table}
 \centering
 \begin{adjustbox}{width=\columnwidth}
\begin{tabular}{@{}llllll@{}}
\toprule
representations & Original & PCA   & KPCA(rbf) & KPCA(sig) & KPCA(lap) \\ \midrule
Word2Vec   & 0.974    & 0.702 & 0.716     & 0.715     &     0.720      \\
Glove  & 0.978    &   0.757       &   0.754    &    0.753       &   0.914   \\ \bottomrule
\end{tabular}
\end{adjustbox}
\caption{Classification accuracy results on male versus female terms.} \label{classification}
\vspace{-15pt}
\end{table}
\subsection{Indirect Bias}
Following \newcite{gonen-goldberg-2019-lipstick-pig}, we build a balanced training set of male and female words using the 5000 most biased words according to the bias in the original representations as described in \cref{subsec:professions}, and then train an RBF-kernel support vector machine (SVM) classifier \cite{scikit-learn} on a random sample of 1000 (training set) of them to predict the gender, and evaluate its generalization on the remaining 4000 (test set). We can perform classification in our corrected RKHS with any SVM kernel
 {$\kk_{\mathrm{svm}}(\neut(\vw), \neut(\vz))$} that can be written in the forms \footnote{Stationary kernels are sometimes written in the form {$\kk(\vw,\vz) = \kk(\langle \vw,\vz \rangle)$} or {$\kk(\vw,\vz) = \kk(||\vw-\vz||^2)$}, i.e., {$\kk_{\mathrm{RBF}}(r) = \exp(-\gamma r^2)$}} {$\kk_{\mathrm{svm}}(\langle \vw, \vz\rangle)$} or {$\kk_{\mathrm{svm}}(||\vw - \vz||^{2})$} since we can use the kernel trick in our corrected RKHS {$\langle \neut(\vw), \neut(\vz) \rangle = \widetilde{\kk}(\vw, \vz)$} to compute the inputs to our SVM kernel, resulting in
\begin{align}
    \kk_{\mathrm{svm}}&(||\vw - \vz||^{2}) \label{eq:svmkern}  \\
    &= \kk_{\mathrm{svm}}(\widetilde{\kk}(\vw, \vw)-2\widetilde{\kk}(\vw, \vz) + \widetilde{\kk}(\vz, \vz)). \nonumber 
\end{align}
It is clear that the RBF kernel is an example of a kernel that follows \cref{eq:svmkern}.

We can see that the bias removal induced by non-linear kernels results in a slightly higher classification accuracy (shown in \cref{classification}) of gendered words for GoogleNews Word2Vec representations \cite{mikolov2013efficient} and a slightly lower classification accuracy for GloVe representations \cite{glove} (with the exception of the Laplace kernel which has a very high classification accuracy). 
Overall for the RBF and the sigmoid kernels there is no improvement in comparison to the linear kernel (PCA), the Laplace kernel seems to have notably worse results than the others, still being able to classify gendered words at a high accuracy of 91.4\% for GloVe representations.

 \subsection{Word Similarity: SimLex-999}
 The quality of a word vector space is traditionally measured by how well it replicates human judgments of word similarity. We use the SimLex-999 benchmark by \citet{hill2015simlex} which provides a ground-truth measure of similarity produced by 500 native English speakers. Similarity scores by our method are computed using Spearman correlation between representation and human judgments are reported. We can observe that the metric corrections only slightly change the Spearman correlation results on SimLex-999 (\cref{tab:simlex}) from the original representation space. We can thus conclude that the representation quality is mostly preserved.

\saveForCR{
\section{Discussion}

We observe that our non-linear extensions of \cite{bolukbasi2016man} report no notable performance differences across a set of benchmarks and statistical tests reported in \cref{sec:experiemtns} designed to quantify bias in word representations. Furthermore the results in \cref{tab:polyglove} (\cref{apdx:poly}) show that gradually increasing the degree of non-linearity has again no significance change in performance for the WEAT \cite{caliskan2017semantics} benchmark. These results provide empirical evidence supporting the \textbf{linear subspace hypothesis} (\cref{eq:hypothesis}) and thus suggest that representing Gender bias in word representations as a linear subspace is a suitable assumption. This result overall has a positive practical impact since as can be seen from the methodology in this paper our non-linear alternative to the method in \cite{bolukbasi2016man} are not only computationally more complex but also partly limited to kernel methods for transfer tasks.
}

\section{Conclusion}
We offer a non-linear extension to the method presented in \citet{bolukbasi2016man} by connecting its bias space construction to PCA and subsequently applying kernel PCA. We contend our extension is natural in the sense that it reduces to
the method of \citet{bolukbasi2016man} in the special
case when we employ a linear kernel and in the non-linear case it preserves all the desired linear properties in the feature space. 
This allows us to provide equivalent constructions of the neutralize and equalize steps presented.\looseness=-1

  \begin{table}
 \centering
 \begin{adjustbox}{width=\columnwidth}
\begin{tabular}{@{}llllll@{}}
\toprule
representations & Original & PCA   & KPCA(rbf) & KPCA(sig) & KPCA(lap) \\ \midrule
Word2Vec   & 0.121    & 0.119 & 0.118     & 0.118     & 0.118     \\
Glove      & 0.302    & 0.298 & 0.298     & 0.298     & 0.305     \\ \bottomrule
\end{tabular}
\end{adjustbox}
\caption{Correlation on SimLex-999 using GoogleNews Word2Vec and GloVe representations. The significance level is {$\alpha = 0.05$} with {$p < 0.001$}.}\label{tab:simlex}
\end{table}

We compare the linear bias mitigation technique to our new kernelized non-linear version across a suite of tasks and datasets. 
We observe that our non-linear extensions of \newcite{bolukbasi2016man} show no notable performance differences across a set of benchmarks designed to quantify gender bias in word representations. Furthermore, the results in \cref{tab:polyglove}(\cref{apdx:poly}) show that gradually increasing the degree of non-linearity has again no significant change in performance for the WEAT \cite{caliskan2017semantics} benchmark. Thus, we provide empirical evidence for the linear subspace hypothesis; our results suggest representing gender bias as a linear subspace is a suitable assumption. We would like to highlight that our results are specific to our proposed kernelized extensions and does not imply that all non-linear variants of \cite{bolukbasi2016man} will yield similar results. There may very well exist a non-linear technique that works better, but we were
unable to find one in this work.

\section{Acknowledgements}

We would like to thank Jennifer C. White for amending several typographical errors in final version of this manuscript.

\bibliographystyle{acl_natbib}
\bibliography{biblio}
\newpage
\appendix 
\onecolumn
\allowdisplaybreaks
\begin{table*}[t!]
\begin{tabular}{@{}llllcllllll@{}}
\toprule
\multirow{2}{*}{Targets} & \multicolumn{2}{l}{Original} & \multicolumn{2}{l}{PCA} & \multicolumn{2}{l}{KPCA (poly-2)} & \multicolumn{2}{l}{KPCA (poly-3)} & \multicolumn{2}{l}{KPCPA (poly-4)} \\ \cmidrule(l){2-11} 
                         & d             & p            & d           & p         & d                & p              & d                & p              & d                & p               \\ \midrule
\multicolumn{11}{l}{    \hspace{3.5cm} Google News Word2Vec \cite{mikolov2013efficient} } \\
\midrule 
Career , Family          & 1.622         & 0.000        & 1.327       & 0.001     & 1.320            & 0.004          & 1.321            & 0.001          & 1.312            & 0.002           \\
Math, Arts               & 0.998         & 0.017        & -0.540      & 0.859     & -0.755           & 0.927          & -0.755           & 0.933          & -0.754           & 0.932           \\
Science , Arts           & 1.159         & 0.005        & 0.288       & 0.281     & 0.271            & 0.312          & 0.272            & 0.305          & 0.272            & 305             \\ \midrule
\multicolumn{11}{l}{    \hspace{5cm} GloVe \cite{glove}} \\
 \midrule
Career , Family          & 1.749         & 0.000        & 1.160       & 0.007     & 1.166           & 0.000          & 1.166           & 0.009          & 1.667           & 0.005          \\
Math, Arts               & 1.162         & 0.007        & 0.144       & 0.389     & 0.096           & 0.429          & 0.097           & 0.421          & 0.097           & 0.432          \\
Science , Arts           & 1.281         & 0.008        & -1.074      & 0.985     & -1.113          & 0.995          & -1.114          & 0.994          & -1.114          & 0.992          \\ \bottomrule
\end{tabular}
\caption{Results for polynomial Kernel Experiments on Glove and Google News representations.} \label{tab:polyglove}
\end{table*}

\section{Polynomial Kernel Results}\label{apdx:poly}

For experimental completeness, we provide direct bias experiments on WEAT using a range of polynomial kernels. 
The results are displayed in \cref{tab:polyglove}. 
The results for the polynomial kernels suggest the same conclusions we arrived at in the main text, i.e., a linear kernel is generally enough.

\end{document}